\documentclass[sts]{imsart}

\RequirePackage{amsthm,amsmath,amsfonts,amssymb}
\RequirePackage[numbers]{natbib}
\RequirePackage[colorlinks,citecolor=blue,urlcolor=blue]{hyperref}
\RequirePackage{graphicx}
\usepackage{booktabs}
\usepackage[inline]{enumitem}
\setlist*[enumerate]{label=(\roman*)}

\startlocaldefs
\numberwithin{equation}{section}
\theoremstyle{plain}
\newtheorem{theorem}{Theorem}
\newtheorem{lemma}[theorem]{Lemma}
\theoremstyle{definition}

\newtheorem{remark}[theorem]{Remark}

\newcommand{\x}{{\bf x}}

\newcommand{\E}{{\mathbb E}}

\newcommand{\X}{{\bf X}}

\newcommand{\btheta}{\boldsymbol{\theta}}

\newcommand{\ben}{\begin{enumerate}}
\newcommand{\een}{\end{enumerate}}
\newcommand{\beq}{\begin{equation}}
\newcommand{\eeq}{\end{equation}}
\newcommand{\bde}{\begin{description}}
\newcommand{\ede}{\end{description}}

\newcommand{\UnifRV}{\mathcal{U}}

\newcommand{\ind}{\stackrel{\mathrm{ind}}{\sim}}

\newcommand{\defeq}{\operatorname{:=}}

\newcommand{\Var}{\mathrm{Var}}

\endlocaldefs

\begin{document}

\begin{frontmatter}
\title{Prediction-Powered Inference with Inverse Probability Weighting}
%\title{A sample article title with some additional note\thanksref{T1}}
\runtitle{PPI with IPW}
%\thankstext{T1}{A sample of additional note to the title.}
\begin{aug}
%%%%%%%%%%%%%%%%%%%%%%%%%%%%%%%%%%%%%%%%%%%%%%%
%% Additional information (such as           %%
%% indicating the corresponding author) can  %%
%% be included in the Acknowledgments        %%
%% section if necessary.                     %%
\author[A]{\fnms{Jyotishka}~\snm{Datta}\ead[label=e1]{jyotishka@vt.edu}}, \and
\author[B]{\fnms{Nicholas G.}~\snm{Polson}\ead[label=e2]{ngp@chicagobooth.edu}}
%%%%%%%%%%%%%%%%%%%%%%%%%%%%%%%%%%%%%%%%%%%%%%
%% Addresses                                %%
%%%%%%%%%%%%%%%%%%%%%%%%%%%%%%%%%%%%%%%%%%%%%%
\address[A]{Department of Statistics, Virginia Tech, Blacksburg, VA. \printead[presep={,\ }]{e1}} 

\address[B]{Booth School of Business, University of Chicago, Chicago, IL. \printead[presep={,\ }]{e2}}
\end{aug}

\begin{abstract}
Prediction-powered inference (PPI) is a recent framework for valid statistical inference with partially labeled data, combining model-based predictions on a large unlabeled set with bias correction from a smaller labeled subset. 
%We show that PPI can be extended to handle informative labeling by replacing its unweighted bias-correction term with an inverse probability weighted (IPW) version, using the classical Horvitz--Thompson or H\'ajek forms. 
Building on existing PPI results under covariate shift, we show that PPI rectification admits a direct design-based interpretation, and that informative labeling can be handled naturally by Horvitz--Thompson and H\'ajek-style corrections.
This connection unites design-based survey sampling ideas with modern prediction-assisted inference, yielding estimators that remain valid when labeling probabilities vary across units. We consider the common setting where the inclusion probabilities are not known but estimated from a correctly specified model. In simulations, the performance of IPW-adjusted PPI with estimated propensities closely matches the known-probability case, retaining both nominal coverage and the variance-reduction benefits of PPI. 
\end{abstract}


\end{frontmatter}
%%%%%%%%%%%%%%%%%%%%%%%%%%%%%%%%%%%%%%%%%%%%%%
%%%% Main text entry area:

\section{Introduction}\label{sec:intro}

% \epigraph{
% ``\textit{Statisticians frequently seek to protect themselves against outrageous fortune by an act of randomization.}"
% ”
% }{\citep{smith1988weight}}

Consider the following prediction problem. We observe a labeled dataset $\{(Y_i, X_i)\}_{i=1}^n$ along with a prediction rule $f(X)$, fitted on the labeled data. We also have a large set of unlabeled covariates $\{\tilde{X}_i\}_{i=1}^N$ with $N \gg n$. Typically the covariates are either synthetic or cheap to produce but the labels or responses are expensive and scarce. Our goal is to construct a valid confidence interval for the estimand $\theta^\star$, e.g., the mean outcome $\theta^\star = \mathbb{E}[Y]$. The central theme of PPI is to leverage the predictions to gain efficiency, while using the labeled data to correct bias and ensure valid coverage. 

%Consider again the prediction problem with a small labeled dataset and a much larger unlabeled set for which model predictions are available, where 

Suppose our parameter of interest is $\theta^\star$, which could be the mean $\E(Y)$, or a specific quantile, or summary from a regression of $Y$ on $\X$ and so on. The prediction-powered inference (PPI) estimator for $\theta^\star$ is:
\[
\hat{\theta}_{\mathrm{PPI}} \;=\; \underbrace{\frac{1}{N} \sum_{i=1}^N f(\tilde{X}_i)}_{\text{predictions on large set}} \;-\; \underbrace{\frac{1}{n} \sum_{i=1}^n \left( f(X_i) - Y_i \right)}_{\text{rectify with labeled error}}
\;=\; \hat{\theta}^f - \hat{\Delta}.
\]
If the predictions are accurate, $\hat{\Delta} \approx 0$ and $\hat{\theta}_{\mathrm{PPI}}$ is close to the average prediction $\frac{1}{N} \sum_{i=1}^N f(\tilde{X}_i)$ but with substantially lower variance. By construction, $\mathbb{E}[\hat{\theta}_{\mathrm{PPI}}] = \theta^\star$, and the two terms in the estimator are independent, allowing the variance for a $95\%$ confidence interval to be obtained by summing the variances of the two parts.

Formally, the PPI estimator combines:
\begin{enumerate*}
    \item a \emph{prediction term} --- the plug-in estimator $\hat{\theta}^f$ computed by applying $f$ to the large unlabeled set, and
    \item a \emph{rectifier term} --- an estimate of the average prediction error over the labeled set,
\end{enumerate*}
to produce the corrected estimate:
\begin{equation}
    \hat{\theta}_{\mathrm{PPI}} \;=\; \hat{\theta}^f - \widehat{\Delta}, 
    \quad \widehat{\Delta} = \frac{1}{n_{\mathrm{lab}}} \sum_{i: R_i=1} \big( f(X_i) - Y_i \big),
    \label{eq:ppi-basic}
\end{equation}
where $R_i$ indicates whether $Y_i$ is observed and $n_{\mathrm{lab}} = \sum_i R_i$.  When the labeled set is a simple random sample (SRS) from the population, $\widehat{\Delta}$ is an unbiased estimator of the mean prediction error, ensuring that $\mathbb{E}[\hat{\theta}_{\mathrm{PPI}}] = \theta^\star$ regardless of the quality of $f$.  \citet{angelopoulos2023prediction} prove that the prediction-powered inference is more powerful than the classical inference based on only labeled data if and only if the size of the unlabeled data ($N$) is substantially larger than the size of the training set (labeled data) $n_{\mathrm{lab}} \defeq n$, and the `model' explains at least some of the variance in $Y$ (i.e., $\Var(f(X_i) - Y_i) < \Var(Y_i)).$

It is well known that lack of reliable labeled data can lead to biased inference with potentially disastrous outcome, and reliable inference in machine learning depends critically on having access to high-quality labeled (or training) data. However, across many scientific fields of enquiry and human enterprise, obtaining labeled or gold-standard data remains challenging as it involves either huge human labour, or costly scientific experiment or securing confidentiality concerns. To allay these roadblocks, researchers have started exploiting deep generative models to cheaply and quickly produce vast amount of labelled data as outputs of deep models, otherwise prohibitive to collect. For example, such predicted `labels' are used in biology for predicting protein structure \citep{jumper2021highly, tunyasuvunakool2021highly}, or climate modeling \citep{hansen2013high}, or for predicting socioeconomic indicators like poverty \citep{steele2017mapping}, deforestation \citep{hansen2013high}, and population densities \citep{robinson2017deep}, based on satellite imagery \citep{rolf2021generalizable}. While such `ML-derived' data allow scientists to draw evidence quickly, they come with further challenges. For one, if predicted labels are used to predict more labels, the biases might amplify. Secondly, the standard tools of classical statistics, such as confidence intervals and p-values that are well-defined in labeled/ gold-standard data lose their meaning in ML-derived data. Thus, a key methodological challenge is to combine predicted data, that are cheap but potentially biased and gold-standard or labeled data, that are reliable but scarce, to arrive at statistically valid inference, without sacrificing the power. 
% lin2023evolutionary (protein)
% jean2016combining, (poverty)
%  ball2017comprehensive (sattelite)

% \paragraph{Prediction-powered inference (PPI):} 

\textbf{Prediction-powered inference (PPI)} \citep{angelopoulos2023prediction, zrnic2024cross, angelopoulos2023ppi++} is a recently proposed framework that fills this gap by leveraging modern prediction algorithms to construct valid statistical inference in partially labeled data settings. The central idea is to use a prediction model to ``fill in'' missing outcomes on a large set of unlabeled units, and then correct for systematic prediction errors using a smaller labeled subset. This approach can dramatically improve efficiency when accurate predictions are available, while still maintaining valid coverage guarantees for the target parameter. 

% To see how this works, suppose we want $\theta = \mathbb{E}[Y]$ but only have $n$ labeled outcomes and $N \gg n$ unlabeled covariates with predictions $\hat{Y}$ from a fitted model. The PPI estimator for the mean is
% \begin{equation}
% \hat{\theta}_{\mathrm{PPI}}
% \;=\;
% \underbrace{\frac{1}{N} \sum_{i=1}^N \hat{Y}_i}_{\text{use predictions at scale}}
% \;-\;
% \underbrace{\frac{1}{n} \sum_{i=1}^n \left( \hat{Y}_i - Y_i \right)}_{\text{rectify with labeled error}}.
% \end{equation}
% The first term is the average prediction over the entire (large) set, while the second term subtracts the \emph{average bias} observed on the labeled data. The variance for a confidence interval combines the variances of these two independent parts in the natural way.

%suppose we wish to estimate a low-dimensional functional $\theta^\star$ of the joint distribution of $(Y, X)$, such as the population mean $\mathbb{E}[Y]$, but we only observe $Y$ for a small labeled set and $X$ for a much larger unlabeled set. Let $f(X)$ be a fitted prediction function for $Y$ based on available labeled data and potentially external information. T

% \paragraph{From PPI to IPW-adjusted PPI.}

\paragraph*{Informative labeling:} In many applications, labeled data are not an SRS but arise from an \emph{informative labeling process}, in which the probability of observing $Y_i$ depends on covariates, $P(R_i = 1) = \xi_i(X_i)$ \citep[see, e.g.,][]{sarndal2003model, little2008weighting}. In this setting, the simple rectifier in \eqref{eq:ppi-basic} could admit some bias. By establishing the connection of PPI estimators with the IPW estimators in survey sampling, we show that this bias can be reduced by replacing the unweighted rectifier with an inverse probability weighted (IPW) version, directly connecting PPI to the Horvitz--Thompson and H\'ajek estimators from survey sampling. 

This work extends the prediction-powered inference (PPI) framework to settings with \emph{informative labeling}, where the probability of observing a label depends on covariates. We show that replacing the standard unweighted bias-correction term in PPI with an inverse probability weighted (IPW) version yields estimators directly connected to the classical Horvitz--Thompson and H\'ajek estimators from survey sampling. This unifies design-based sampling ideas with modern prediction-assisted inference, ensuring validity under unequal labeling probabilities. We also establish a novel connection between PPI and importance sampling via the vertical likelihood representation: the PPI correction step parallels higher-order quadrature bias reduction (e.g., trapezoidal over rectangular rules), clarifying when and why PPI can yield substantial variance reduction without sacrificing coverage.

\noindent \textbf{Relation to existing PPI work.} Prediction-powered inference under covariate shift and unequal labeling probabilities has already been considered in the original PPI framework of Angelopoulos et al. (2023) \citep{angelopoulos2023prediction}, where inverse probability weighting appears as a special case of their general correction scheme (see Section 4.2 and Corollary 1 therein). In particular, their results already imply a Horvitz--Thompson-type correction that yields valid inference under known inclusion probabilities.

Our contribution is not to reintroduce inverse probability weighting into PPI, but rather to make explicit the precise connection between PPI rectification and classical design-based estimators, and to develop this connection systematically. By viewing the PPI bias-correction term as a Horvitz--Thompson or H\'ajek estimator applied to prediction residuals, we place PPI squarely within the survey sampling literature, clarify the role of normalization and ratio estimation, and highlight practical variants that are natural from a design-based perspective but have received little attention in the PPI literature.

The outline of the paper is as follows. In \S\ref{sec:ppi-ipw}, we survey existing IPW estimators and develop inverse probability weighted extensions of prediction-powered inference, linking the PPI rectifier to the H\'ajek and Horvitz--Thompson estimators and discuss the use of estimated inclusion probabilities. We also discuss the connections between PPI and vertical likelihood ideas as well as cross-PPI \citep{zrnic2024cross} with the binning-smoothing approach of \citet{ghosh2015weak} in \S \ref{sec:bin}. 
In \S\ref{sec:sim}, we present numerical experiments with synthetic and real data comparing the proposed estimators to their unweighted counterparts and to classical design-based estimators under informative labeling. Finally, in \S\ref{sec:diss}, we provide a brief discussion of the main findings. Code to reproduce all analyses is available at \url{https://github.com/dattahub/ppi_ipw}.

\section{Incorporating IPW into Prediction-Powered Inference}\label{sec:ppi-ipw}

\subsection{Horvitz--Thompson and H\'ajek Estimators}
While inverse probability weighting appears implicitly in earlier PPI formulations, adopting an explicit design-based perspective reveals additional structure. In particular, it shows that the standard PPI rectifier coincides exactly with a H\'ajek ratio estimator under simple random sampling, and suggests natural IPW generalizations under informative labeling.
\textcolor{black}{We start with a brief description of two of the most widely used estimators in survey sampling are the Horvitz--Thompson (HT) estimator \citep{horvitz1952generalization}\footnote{Also called the Narain--Horvitz--Thompson estimator after \citet{narain1951sampling}; see \citet{rao1999some, chauvet2014note}.} and the H\'ajek estimator \citep{hajek1971comment}.  
Throughout, we focus on the finite-population mean $\theta_N = N^{-1} \sum_{i=1}^N Y_i$, the design-based target in survey sampling, while noting that the superpopulation mean $\theta_f = \mathbb{E}_f[Y]$ under a data-generating model can be treated analogously with the same estimators.}

Consider a finite population $U = \{1, 2, \dots, N\}$ with associated values $Y_k$ for each unit $k \in U$.  A sample $s \subset U$ is drawn according to a sampling design with inclusion probabilities $p_k = P(k \in s)$.  The Horvitz--Thompson and the H\'ajek ratio estimator for the population mean $\theta = N^{-1} \sum_{k \in U} Y_k$ are:
\begin{equation}
    \hat{\theta}_{\mathrm{HT}} = \frac{1}{N} \sum_{k \in s} \frac{Y_k}{p_k}. \qquad \hat{\theta}_{\mathrm{H\acute{a}jek}} = \frac{\sum_{k \in s} Y_k / p_k}{\sum_{k \in s} 1 / p_k}.
    \label{eq:ht}
\end{equation}
The H\'ajek estimator replaces the known $N$ in the denominator with its design-based estimate $\hat{N} = \sum_{k \in s} 1/p_k$.  While slightly biased in finite samples, it often has lower variance than HT and does not require the population size $N$ to be known. In missing data terms \citep[see][]{khan2023adaptive}, with $R_k$ indicating observation of $Y_k$, the Horvitz--Thompson and H\'ajek estimator are:
\begin{align}
    \hat{S} = \sum_{k=1}^{N}\frac{Y_k R_k}{p_k} &, \quad \hat{S} = \sum_{k=1}^{N}\frac{R_k}{p_k}, \quad
    \hat{\theta}_{HT} = \frac{\hat{S}}{N}, \quad     \hat{\theta}_{\text{H\'ajek}} = \frac{\hat{S}}{\hat{N}}. \label{eq:ipw-2}
\end{align}

\textcolor{black}{We note here that the idea of leveraging available labeled data to obtain unbiased predictions or valid inference for unlabeled units has deep roots in the statistical literature. A notable example is the ratio estimators in \eqref{eq:ht}, which connects directly to PPI’s bias-correction term in the finite-population setting and, under certain conditions, can improve the residual adjustment in PPI. Other methods, such as partial least squares, share the related philosophy of using observed labels to uncover predictive structure in covariates, although their strategies for combining labeled and unlabeled information differ.}

\textcolor{black}{\citet{royall1982balanced} demonstrated that the ratio estimator arises naturally when the labeled and unlabeled observations are jointly modeled as independent normal random variables with means proportional to a known auxiliary variable $x_i$, under a diffuse prior on the regression coefficient. Specifically, the ratio estimator coincides with the Bayes posterior predictive mean under the heteroscedastic regression model $E(Y_i \mid \beta) = \beta x_i$, $\mathrm{var}(Y_i \mid \beta) = \sigma^2 x_i$, with a flat prior on $\beta$. \citet{smith1988weight} also argued in favor of using inverse-selection-probability weights as the basic device for estimating population totals when sampling is unequal. Another example of a model-assisted framework is the generalized regression (GREG) estimator of \citet{cassel1976some} who adjusts the Horvitz--Thompson (HT) estimator using auxiliary information to improve efficiency. It is defined as
\[
\hat{Y}_{\mathrm{\small GR}} = \hat{Y}_{\mathrm{\small HT}} + \hat{\beta}^\top \left( X - \hat{X}_{\mathrm{\small HT}} \right),
\]
where $X = \sum_{i \in U} x_i$ is the known finite-population total of auxiliary variables $x_i$, $\hat{X}_{\mathrm{\small HT}} = \sum_{k\in s} x_k / p_k$ is the HT estimate of $X$, and $\hat{\beta}$ is estimated from the sample via regression of $y_i$ on $x_i$. The adjustment term corrects for discrepancies between the known $X$ and its HT estimate, leveraging the correlation between $y_i$ and $x_i$ to reduce variance and improve accuracy, without altering the unbiasedness of the HT estimator.}

\textcolor{black}{Like the Horvitz--Thompson estimator in survey sampling, the PPI estimator achieves design-based unbiasedness: $E[\hat{\theta}_{\mathrm{PPI}}] = \theta^* = N^{-1} \sum_{i=1}^N Y_i$, where the expectation is taken over the randomness in the sampling mechanism rather than any assumed data-generating process. This design-based property ensures that the estimator remains unbiased for the finite-population parameter $N^{-1} \sum_{i=1}^N Y_i$ regardless of whether the prediction model $f(\cdot)$ correctly captures the relationship between covariates and outcomes. Recently, \citet{datta2025inverse} investigated the inverse probability weighting estimators including the H\'ajek, HT, a Bayesian estimator due to \citet{li2010bayesian} and a binning-smoothing estimator due to \citet{ghosh2015weak} in the context of an weak paradox due to \citet{wasserman2004bayesian}.}

\textcolor{black}{Next, we show how these IPW formulations can be used to derive a bias-correction component, and propose IPW-adjusted PPI estimators, with the HT and H\'ajek forms in \eqref{eq:ht} corresponding directly to the weighted rectifiers in \eqref{eq:ppi-ht}. \citet{angelopoulos2023ppi++} provides a semiparametric missing-data and AIPW perspective of PPI++ using the known propensity score, or probability of missingness $n_{\mathrm{lab}}/N$ in the MCAR situation. Our contribution ties PPI directly to Horvitz--Thompson and H\'ajek estimation and studies informative labeling and cross-fitted implementations.}

\subsection{IPW-adjusted PPI estimators}
We begin with the prediction-powered inference (PPI) estimator for a finite population mean $\theta^* = N^{-1} \sum_{i=1}^N Y_i$,
\begin{equation}
\hat{\theta}_{\mathrm{PPI}}
\;=\;
\underbrace{\frac{1}{N} \sum_{i=1}^N \hat{Y}_i}_{\text{prediction term}}
\;-\;
\underbrace{\frac{1}{n_{\mathrm{lab}}} \sum_{i: R_i=1} \left( \hat{Y}_i - Y_i \right)}_{\text{rectifier term}},
\label{eq:ppi-basic-mapping}
\end{equation}
where $R_i \in \{0,1\}$ indicates whether $Y_i$ is observed, $n_{\mathrm{lab}} = \sum_{i=1}^N R_i$, and $\hat{Y}_i = f(X_i)$ is a fitted prediction from covariates $X_i$.  When $R_i$ is generated by simple random sampling, the rectifier term is an unbiased estimate of the mean prediction error, ensuring that $\mathbb{E}[\hat{\theta}_{\mathrm{PPI}}] = \theta^*$. Letting $\xi_i = P(R_i = 1)$ denote the inclusion probability, the Horvitz--Thompson and H\'ajek estimators for the population mean are: 
\begin{equation}\label{eq:ht-hajek}
\hat{\theta}_{\mathrm{HT}} 
= \frac{1}{N} \sum_{i=1}^N \frac{R_i Y_i}{\xi_i}, 
\qquad 
\hat{\theta}_{\mathrm{H\acute{a}jek}} 
= \frac{\sum_{i=1}^N \frac{R_i Y_i}{\xi_i}}{\sum_{i=1}^N \frac{R_i}{\xi_i}}.
\end{equation}
It is easy to see that if $\xi_i \equiv \xi$ is constant (simple random sampling), %$$(1/n_{\mathrm{lab}})\sum_{i:\,R_i=1}(\hat{Y}_i - Y_i)=\big(\sum_{i=1}^N R_i(\hat{Y}_i - Y_i)/\xi\big)\ /\ \big(\sum_{i=1}^N R_i/\xi\big)$$, 
\begin{equation}
\frac{1}{n_{\mathrm{lab}}} \sum_{i: R_i=1} 
\big( \hat{Y}_i - Y_i \big)
= 
\frac{\sum_{i=1}^N \frac{R_i (\hat{Y}_i - Y_i)}{\xi}}
{\sum_{i=1}^N \frac{R_i}{\xi}},
\end{equation}
which is exactly the H\'ajek form applied to the residual $e_i = \hat{Y}_i - Y_i$.  
Hence, under simple random sampling, PPI is a \emph{H\'ajek ratio estimator} of the mean prediction error, subtracted from the full‐population mean of $\hat{Y}_i$.

\paragraph*{Informative labeling.} Now, consider the situation where the probability of labeling depends on available covariates, i.e., $P(R_i = 1 \mid X_i) = \xi_i$ with $\xi_i$ not constant. Such a scheme could be realistic in studies where data collection or annotation is more likely for certain groups: e.g., younger participants may be overrepresented due to recruitment convenience or targeted study designs. In such cases, the unweighted rectifier in \eqref{eq:ppi-basic-mapping} is biased.  
In this case, we can replace it with an \emph{inverse probability weighted} (IPW) estimator of the mean prediction error.  Two natural choices, paralleling the Horvitz--Thompson and H\'ajek estimators in survey sampling, are:
\begin{align}
\delta_{\mathrm{HT}} &= \frac{1}{N} \sum_{i=1}^N \frac{R_i}{\xi_i} \, e_i,
\quad \delta_{\mathrm{H\acute{a}jek}} = 
\frac{\sum_{i=1}^N \frac{R_i}{\xi_i} \, e_i}{\sum_{i=1}^N \frac{R_i}{\xi_i}},
\label{eq:ppi-hajek-rectifier}
\end{align}
where $e_i = \hat{Y}_i - Y_i$ denotes the prediction residual for unit $i$. Replacing the rectifier in \eqref{eq:ppi-basic-mapping} with \eqref{eq:ppi-hajek-rectifier} yields:
\begin{align*}
\hat{\theta}_{\mathrm{PPI,HT}} 
&= \frac{1}{N} \sum_{i=1}^N \hat{Y}_i \;-\; \delta_{\mathrm{HT}}, \\
\hat{\theta}_{\mathrm{PPI,H\acute{a}jek}} 
& = \frac{1}{N} \sum_{i=1}^N \hat{Y}_i \;-\; \delta_{\mathrm{H\acute{a}jek}}.
\end{align*}
The HT form is design-unbiased given correct $\xi_i$, while the H\'ajek form is approximately unbiased but often exhibits lower variance.
% \paragraph{Estimated propensities as a robustness check.}
The inclusion probabilities $\xi_i$ may be known exactly in designed experiments or probability samples, but in observational settings with informative missingness, they must often be estimated from the data by fitting a model for $R_i \mid X_i$, for example via logistic regression. In our simulations, using $\hat{\xi}_i$ in place of $\xi_i$ had negligible impact on bias, coverage, or efficiency when the propensity model was correctly specified. This serves as a robustness check, supporting the use of IPW-adjusted PPI in realistic settings where $\xi_i$ must be estimated. Formally, using the Horvitz--Thompson form,
\begin{equation}
    \widehat{\Delta}_{\mathrm{HT}} 
    = \frac{1}{N} \sum_{i=1}^N \frac{R_i}{\hat{\xi}_i} \, \big( f(X_i) - Y_i \big),
    \label{eq:ppi-ht}
\end{equation}
or the H\'ajek form,
\begin{equation}
    \widehat{\Delta}_{\mathrm{H\acute{a}jek}} 
    = \frac{\sum_{i=1}^N \frac{R_i}{\hat{\xi}_i} \, \big( f(X_i) - Y_i \big)}{\sum_{i=1}^N \frac{R_i}{\hat{\xi}_i}},
    \label{eq:ppi-hajek}
\end{equation}
yields an \emph{IPW-adjusted PPI estimator} that remains unbiased (or approximately unbiased in the H\'ajek case) under informative labeling.

\begin{remark}
    \textcolor{black}{In the missing-data taxonomy, this setting corresponds to the \emph{missing at random} (MAR) case, i.e., $R_i \perp Y_i \mid X_i$, and in our numerical experiments and real data analysis, we assume that the labeling probability $\xi_i$ depends only on observed covariates (e.g., age). Under MAR, inverse probability weighting (IPW) and its PPI analogue yield unbiased estimates of the finite-population mean. The usual set-up for PPI is the missing completely at random (MCAR); $R_i \perp (Y_i, X_i)$. Under MCAR, the PPI residual term coincides with the Horvitz--Thompson correction. In our simulations, If, instead, $R_i$ depended directly on $Y_i$ even after conditioning on $X_i$, the mechanism would be \emph{missing not at random} (MNAR, or nonignorable missingness).}
\end{remark}

\subsection{Variance estimation and confidence intervals.}
The variance formulas presented below are derived under independent 
Bernoulli labeling, $R_i \ind \text{Bernoulli}(\xi_i)$, and assume 
that the prediction model $f$ is trained on data independent of the 
labeled sample \citep{angelopoulos2023prediction}, so that the prediction 
term $\frac{1}{N}\sum_{i=1}^N \hat{Y}_i$ and the rectifier 
$\delta_{\text{HT}}$ are independent. Under this assumption, the 
variance of the prediction term is $O(N^{-1})$ and negligible when 
$N \gg n$, yielding
\begin{equation}
\text{Var}(\hat{\theta}_{\text{PPI,HT}}) = \text{Var}(\delta_{\text{HT}}) + O(N^{-1}).
\end{equation}
When $f$ is instead trained on the labeled data, the independence 
assumption fails and the cross-fitting approach of 
\S\ref{sec:bin} provides a remedy. For the HT-adjusted PPI estimator $\hat{\theta}_{\text{PPI,HT}}$, 
we can approximate the variance as:
\begin{equation}
\text{Var}(\hat{\theta}_{\text{PPI,HT}}) \approx \text{Var}(\delta_{\text{HT}}),
\end{equation}
where
\[
\delta_{\text{HT}}=\frac{1}{N}\sum_{i=1}^N \frac{R_i}{\xi_i}e_i,
\qquad e_i=\hat Y_i-Y_i.
\]
Under independent Bernoulli labeling, substituting estimated 
propensities $\hat{\xi}_i$ for the unknown $\xi_i$ following standard 
practice \citep{sarndal1992model}, this variance can be estimated by
\begin{equation}
\widehat{\text{Var}}(\delta_{\text{HT}})=\frac{1}{N^2}\sum_{i=1}^N 
\frac{1-\hat{\xi}_i}{\hat{\xi}_i^2} R_i e_i^2.
\end{equation}
A 95\% confidence interval is then
\begin{equation}
\hat{\theta}_{\text{PPI,HT}} \pm 1.96\sqrt{\widehat{\text{Var}}
(\delta_{\text{HT}})}.
\end{equation}
Note that for more complex survey designs with non-independent 
sampling, additional terms involving joint inclusion probabilities 
$\pi_{ij}$ would be required \citep{sarndal1992model}. 

For the H\'ajek-adjusted PPI estimator $\hat{\theta}_{\text{PPI,H\'ajek}}$, the variance estimation is more complex due to the ratio structure. Writing $A = \sum_{i=1}^N \frac{R_i}{\xi_i} e_i$ and $B = \sum_{i=1}^N \frac{R_i}{\xi_i}$, a standard delta method expansion of $\delta_{\text{H\'ajek}} = A/B$ around $(\mathbb{E}[A], \mathbb{E}[B])$ shows that the bias is of order $O(n^{-1})$ under independent Bernoulli sampling. The same linearization yields a conservative variance estimator, standard in 
survey sampling \citep{sarndal1992model, datta2025inverse}: 
\begin{equation}
\widehat{\text{Var}}(\delta_{\text{H\'ajek}}) = \frac{1}{\left(
\sum_{i=1}^N \frac{R_i}{\hat{\xi}_i}\right)^2} \sum_{i=1}^N 
\frac{1-\hat{\xi}_i}{\hat{\xi}_i^2} R_i (e_i - 
\hat{\delta}_{\text{H\'ajek}})^2,
\label{eq:hajek-var}
\end{equation}
which accounts for the centering inherent in the ratio estimator (where $\hat{\delta}_{\text{H\'ajek}}$ is evaluated at the estimated propensities $\hat{\xi}_i$). Under the same assumption that $f$ is trained on data independent of 
the labeled sample \citep{angelopoulos2023prediction}, an approximate 95\% 
confidence interval is:
\begin{equation}
\hat{\theta}_{\text{PPI,H\'ajek}} \pm 1.96 \sqrt{\widehat{\text{Var}}
(\delta_{\text{H\'ajek}})}.
\end{equation}
When inclusion probabilities $\xi_i$ must be estimated, additional 
uncertainty arises. The variance formulas above condition on the 
estimated propensities $\hat{\xi}_i$ and do not account for propensity 
estimation uncertainty. In our simulations (\S \ref{sec:sim}), we 
found that treating $\hat{\xi}_i$ as fixed in the variance 
calculations -- a standard practice in survey sampling when propensity 
models are correctly specified -- yields approximately valid coverage. 
For more conservative inference that fully accounts for propensity 
estimation uncertainty, bootstrap methods that resample both the 
propensity estimation and the IPW correction can be employed. Specifically, one would:
\begin{enumerate}
\item Draw a bootstrap sample $s^*$ from the labeled units with 
    probabilities proportional to $1/\hat{\xi}_i$
\item Re-estimate propensities $\hat{\xi}_i^*$ on $s^*$
\item Compute $\hat{\theta}_{\text{PPI}}^*$ using the bootstrap 
    propensities
\item Repeat $B$ times to obtain the bootstrap distribution
\end{enumerate}
The empirical quantiles of $\{\hat{\theta}_{\text{PPI}}^*\}_{b=1}^B$ 
then provide approximately valid confidence intervals, under correct 
propensity model specification, that account for both sampling 
variability and propensity estimation uncertainty.

\subsection{Cross-fitting and Binning--Smoothing}\label{sec:bin}

\citet{zrnic2024cross} propose a cross-fitting variant of prediction-powered inference that trains the model $f$ out of fold and debiases the imputed labels, addressing the failure that can occur if one uses the same labeled data to both train 
the model and estimate the PPI correction. The labeled data are partitioned into $K$ folds $\{\mathcal{I}_k\}_{k=1}^K$. For each fold $k$, a prediction function $f^{(-k)}$ is fitted using all labeled data except those in $\mathcal{I}_k$, and then applied to produce predictions for both the unlabeled set and the held-out labeled fold. The PPI estimator is computed fold-by-fold and averaged:
\begin{align*}
    \hat{\theta}_{\mathrm{cross\text{-}PPI}} \; & =\; \frac{1}{K} \sum_{k=1}^K \bigg\{ \frac{1}{N} \sum_{i=1}^N f^{(-k)}(X_i) \\ & - \frac{1}{n_k} \sum_{i \in \mathcal{I}_k} \big( f^{(-k)}(X_i) - Y_i \big) \bigg\},
\end{align*}
where $n_k = |\mathcal{I}_k|$. Cross-fitting removes the optimistic bias in the bias-correction term that can occur if the same data are reused for training and evaluation.

When the labeling probabilities $\xi_i$ are known or estimated, the bias-correction term can be further stabilized by incorporating design-based ideas from the survey sampling literature. \citet{ghosh2015weak} proposed a \emph{binning-and-smoothing} estimator for the Horvitz--Thompson mean: first, group the $\xi_i$ into $B$ bins $[a_1, a_2), \dots, [a_B, a_{B+1}]$ so that probabilities within a bin are approximately equal; second, replace each $\xi_i$ in bin $b$ by the bin midpoint $p_b$; and third, compute the HT estimate within each bin and average across bins:
\[
\hat{\theta}_{\mathrm{BS}} = \sum_{b=1}^{B} \frac{n_b}{n} \cdot \frac{1}{n_b} \sum_{i \in \mathcal{B}_b} \frac{R_i Y_i}{p_b},
\]
where $n_b$ is the number of labeled units in bin $b$ and $\mathcal{B}_b$ is the set of labeled indices in that bin. This ``coarse graining'' of $\xi_i$ values could potentially reduce variance by borrowing strength across units with similar inclusion probabilities, while maintaining design-unbiasedness under mild regularity conditions \citep{datta2025inverse}.

% As discussed before, \citet{datta2025quantile} connected these ideas to the \emph{vertical likelihood} framework, showing that suitable smoothing of the likelihood-ordinate function $\Lambda(s)$ can yield higher-order convergence rates in numerical integration problems (e.g., $O(n^{-4})$ using the Yakowitz trapezoid rule). In the context of PPI with informative labeling, cross-fitting mitigates overfitting bias, while binning-and-smoothing might an analogous role to higher-order quadrature rules in reducing variance. %Together, these strategies offer a robust and efficient approach to valid inference when labeled data are scarce and nonrandomly selected.

\subsection{Connecting Vertical Likelihood with PPI}

In this section, we shift perspective from design-based inference to a Bayesian integration problem, showing an analogy between PPI rectification and higher-order quadrature methods in importance sampling. While somewhat tangential to our main development, this connection provides an alternative lens through which to understand the variance-reduction benefits of PPI.

An alternative way to interpret the link between sampling schemes and numerical integration is through the lens of a missing-data problem. Suppose our aim is to evaluate  
\begin{equation}
    		\theta = \E_{F}{L(\btheta)} = \int_{\chi} L(\btheta) dF(\btheta). \label{eq:psi}
\end{equation}
Given a sample $(y_1,\dots,y_n)$ drawn either from the target density $f$ or from a proposal density $g$, the basic IS estimate is  $\hat{\theta}_{\mathrm{IS}} = \frac{1}{n} \sum_{i=1}^n l(y_i) f(y_i)/g(y_i)$, which reduces to the simple empirical mean $\hat{\theta} = n^{-1} \sum_{i=1}^n l(y_i)$ when $f = g$. In that case, the Law of Large Numbers ensures convergence to $\theta$, and the Central Limit Theorem implies that the mean squared error converges to 0 at rate $O(n^{-1})$. Prior work has shown that replacing the empirical average with a Riemann-sum approximation can yield marked improvements in stability and convergence \citep{philippe1997processing, philippe2001riemann, yakowitz1978weighted}. More recently, \citet{datta2025quantile} demonstrated that combining Riemann-sum estimators with nested sampling can accelerate convergence further---up to $O(n^{-4})$ in some Bayesian marginal likelihood problems. \citet{datta2025quantile} proposed the Quantile Importance Sampling (QIS) that exploits the Lorenz identity,
\[
\theta = \int_0^1 \Lambda(s) \, ds, \quad \Lambda(s) = \sup\{t \in \mathbb{R} : F_L(t) \le 1-s\},
\]
where $F_L$ is the cumulative distribution function of the likelihood ordinate under the prior.  
This vertical likelihood representation reduces a $p$-dimensional integral to a univariate one over $[0,1]$, and applying the Yakowitz trapezoid rule to the ordered quantiles of $\Lambda(s)$ yields an $O(n^{-4})$ convergence rate under mild smoothness conditions. We present here a lemma from \citet{datta2025quantile}. For details on vertical likelihood and an unifying perspective for several sampling-based strategies see \citet{datta2025quantile}. 
\begin{lemma}\label{prop:1}
Let the evidence be written as $$\theta \equiv \int_{\chi} L(\x) \, dF(\x) = \int_0^1 \Lambda(s) \, ds$$. Suppose $\Lambda(s)$ has a continuous first derivative and a bounded second derivative $\Lambda''(s)$ on the unit interval. Define $U_{(0)} \equiv 0$, $U_{(n+1)} \equiv 1$, and let $\{U_{(i)}\}_{i=1}^n$ denote the order statistics from $n$ independent $\UnifRV(0,1)$ draws (so $U_{(i)} \ge U_{(i-1)}$ for $i = 1, \ldots, n+1$). The QIS estimator is
\[
\hat{\theta}_{QIS} = \frac12 \sum_{i=1}^{n+1} (U_{(i)} - U_{(i-1)}) \left[ \Lambda(U_{(i-1)}) + \Lambda(U_{(i)}) \right].
\]
Then, for some constant $M > 0$, $\E\left[(\theta - \hat{\theta}_{QIS})^2\right] \leq \frac{M}{n^4},$ for all $n \geq 1.$
\end{lemma}
% The estimand in \eqref{eq:psi} can be regarded as the limit of the finite average $\theta_n = n^{-1} \sum_{i=1}^n l(\theta_i)$ as $n \to \infty$, where $\theta_i \sim F(\cdot)$. For a sufficiently large $N$, the finite-population mean $\theta_N = N^{-1} \sum_{i=1}^N l(\theta_i)$ can approximate $\theta$ to any desired accuracy. If we only observe a random subset of size $n \ll N$, with sampling probabilities $\pi_i$ attached to the observed $l(\theta_i)$, then estimating $\theta$ reduces to the familiar problem of estimating $\theta_N$ from $\theta_n$. In this view, the standard importance sampling (IS) estimator plays a role analogous to the Horvitz--Thompson (HT) estimator, $\mathbb{E}_{\pi}[\sum_{i=1}^n l(\theta_i) / \pi_i] = \theta$. As with HT, poor choice of proposal distribution $g(\cdot)$ in IS can inflate variance, even though bias may vanish asymptotically. This connection has motivated a range of modified IS methods designed to improve stability.  

The connection between IS and PPI can be seen as follows. In IS, one typically observes the likelihood ordinates at only a subset of points in $[0,1]$, while the remainder of the interval is ``unlabeled.'' The bias from this partial observation can be corrected using the available $(Y, X)$ pairs. In numerical integration terms, this debiasing step corresponds to the trapezoid rule, whereas classical Monte Carlo integration corresponds to the rectangular rule; the trapezoid rule can achieve convergence rates up to $O(n^{-4})$. In other words, the gain from nested sampling over classical IS is analogous to the improvement that PPI offers over the classical sample mean estimator.

\section{Numerical Experiments}\label{sec:sim}

\subsection{Real Data: Predicting BMI with NHANES Data}\label{sec:nhanes}
Throughout our numerical experiments, we compare five estimators:
\begin{itemize}
\item \textbf{Classic:} The sample mean of labeled data only, $\bar{Y}_{\text{lab}} = n_{\text{lab}}^{-1} \sum_{i: R_i=1} Y_i$
\item \textbf{Horvitz--Thompson (HT):} The IPW estimator using estimated propensities $\hat{\xi}_i$, as defined in \eqref{eq:ht-hajek}
\item \textbf{H\'ajek:} The ratio estimator using estimated propensities $\hat{\xi}_i$, as defined in Equation~\eqref{eq:ht-hajek}
\item \textbf{PPI (unweighted):} The standard PPI estimator assuming MCAR, as defined in Equation~\eqref{eq:ppi-basic}
\item \textbf{PPI (weighted):} The H\'ajek-adjusted PPI estimator for informative labeling, using the rectifier in Equation~\eqref{eq:ppi-hajek}
\end{itemize}
To illustrate their performance on real data, we use the National Health and Nutrition Examination Survey (\texttt{NHANES}) 2013--2014 dataset from CDC \citep{cdc_nhanes}, a nationally representative health and nutrition survey conducted in the United States, retrieved using the \texttt{nhanesA} R package. The NHANES database combines interviews and physical examinations to collect a wide range of demographic, socioeconomic, dietary, and health-related variables. Here, we treat the \texttt{NHANES} dataset as the finite population and the mean body mass index (\texttt{BMI}) as the estimand. In the pre-processing stage, we perform an inner join between the demographics and the body measurements datasets, and remove all rows with missing values, to avoid the vagaries of imputation. We introduce an \emph{informative labeling} mechanism by making the probability of a unit being labeled depend on age, and treat this probability as the true inclusion probability $\xi_i$. In particular, we take $\xi_i = \sigma(3 - 0.05 \times \mathrm{Age}_i)$, where $\sigma = 1 / \{1 + \exp(-x)\}$ is the logistic sigmoid, and the labels $R_i$'s are drawn as $\mathrm{Bernoulli}(\xi_i)$, for all $i = 1, \ldots, n$. We fit a linear regression of \texttt{BMI} on age, gender, waist circumference, upper arm circumference, upper leg length, and race using the labeled subset, and use the fitted model to generate predictions for all units. 
\begin{figure}[ht!]
    \centering
    \includegraphics[width=0.9\linewidth]{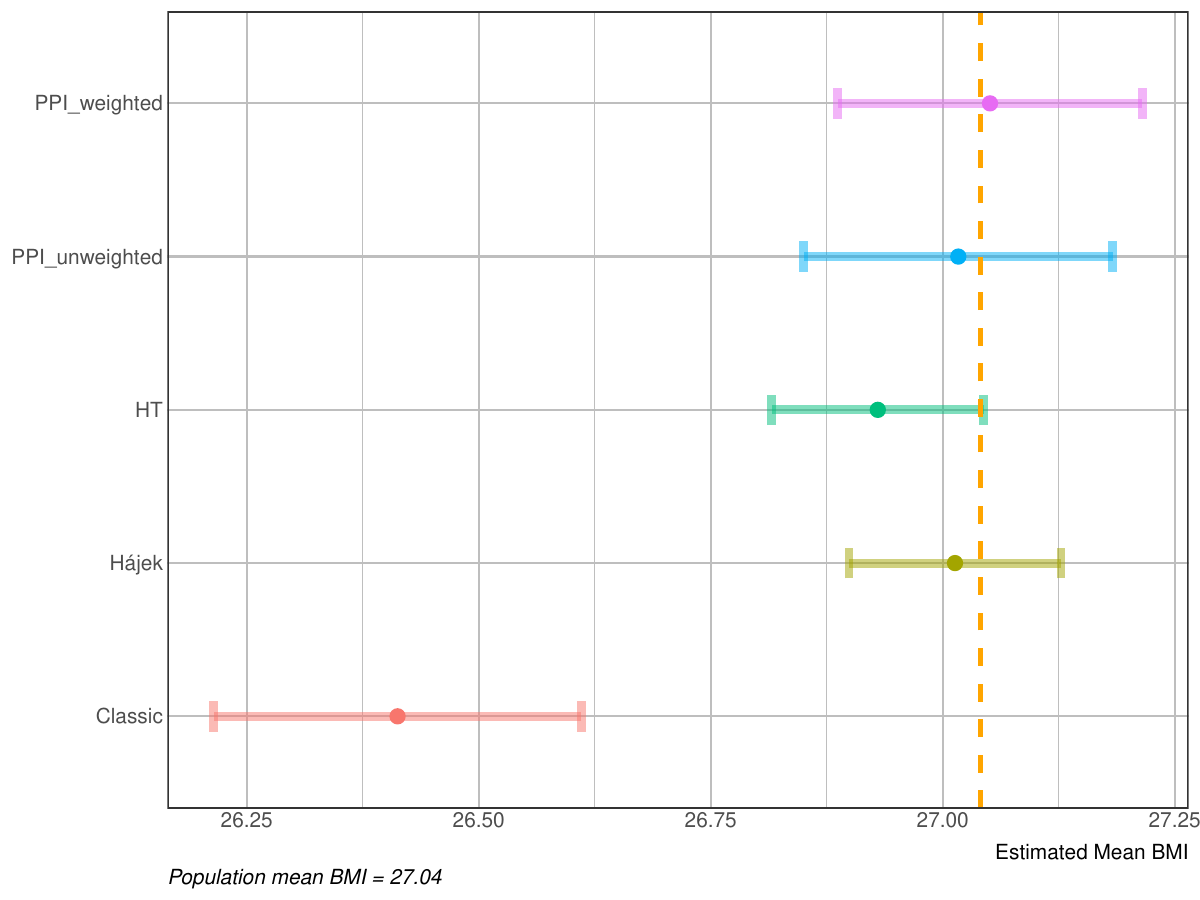}
    \caption{NHANES example: point estimates and 95\% confidence intervals for the Classic mean, Horvitz--Thompson (HT), H\'ajek, and PPI rectifiers with and without weighting, under informative labeling depending on age. The dashed line marks the population mean BMI for the NHANES dataset after omitting the missing values across columns.}
    \label{fig:nhanes_ppi}
\end{figure}
Figure \ref{fig:nhanes_ppi} displays point estimates and 95\% confidence intervals for the mean BMI using the Classic mean, Horvitz--Thompson (HT), H\'ajek, and PPI rectifiers with and without weighting. The dashed line marks the true population mean BMI of 27.04 for the NHANES dataset (after omitting the missing values in the covariates). The classic estimator is biased under the informative labeling, while the Horvitz--Thompson estimator partially corrects this bias. The H\'ajek estimator and both weighted and unweighted PPI achieve substantially lower bias, with the weighted PPI nearly unbiased. The empirical biases were $-0.628$ (Classic mean), $-0.111$ (Horvitz--Thompson), $-0.0277$ (H\'ajek), $-0.0241$ (unweighted PPI), and $0.0101$ (weighted PPI), while the corresponding 95\% confidence interval widths were $0.396$, $0.229$, $0.229$, $0.333$, and $0.328$, respectively. 

This example illustrates the bias-variance tradeoff inherent in these methods. The Horvitz--Thompson estimator achieves the narrowest confidence intervals but retains moderate bias from prediction errors. The H\'ajek estimator substantially reduces this bias through ratio estimation while maintaining the same narrow interval width, making it highly competitive. The PPI methods achieve further bias reduction, with the weighted PPI approach nearly eliminating bias while accepting a modest increase in interval width. Overall, the H\'ajek estimator, unweighted PPI, and weighted PPI all perform well, each offering different points on the bias-variance frontier. We also note that the relative performance could change for larger datasets or under various phenomena such as covariate shift or distribution shift \citep{angelopoulos2023prediction}.

\subsection{Synthetic Data}\label{sec:synthetic}
Similar to Section \ref{sec:nhanes}, we evaluated 5 estimators of a binary population mean under an informative labeling design:
(i) the classic sample mean of labeled data;
(ii) the Horvitz--Thompson (HT) estimator with $\hat{\xi}_i$;
(iii) the H\'ajek estimator with $\hat{\xi}_i$;
(iv) the prediction-powered inference (PPI) estimator with the unweighted rectifier from \citet{angelopoulos2023prediction}; and
(v) a weighted PPI variant using a H\'ajek-style rectifier with $\hat{\xi}_i$. Since our approach estimates the propensities $\xi_i$ from the observed data, any standard IPW estimator, such as survey-weighted Horvitz--Thompson or H\'ajek, can be directly applied to the labeled subset as a baseline for comparison.
% \paragraph{Propensity estimation.}
For each simulated dataset, we estimated $\xi_i$ by fitting a logistic regression of $R_i$ on the covariate $X_i$ and used the fitted probabilities $\hat{\xi}_i$ in place of the truth for all IPW calculations.

A super-population of $N=500$ units was generated from a logistic model with a single covariate $X_i \sim N(0,1)$ and $Y_i \sim \mathrm{Bernoulli}(\mathrm{logit}^{-1}(X_i))$. Inclusion probabilities $\xi_i$ were set to $\mathrm{logit}^{-1}(0.5 X_i)$, and labels $R_i \sim \mathrm{Bernoulli}(\xi_i)$ were drawn independently. Predictions $\hat{Y}_i$ were generated from $\mathrm{logit}^{-1}(X_i + \varepsilon_i)$ with $\varepsilon_i \sim N(0,0.5^2)$.  

For each of 200 replicates, we computed all five estimators and formed 95\% confidence intervals. We summarized bias, mean interval width, and empirical coverage, and plotted confidence intervals from the first 10 replicates alongside their widths.

\begin{figure}[h!]
    \centering
    \includegraphics[width=\linewidth]{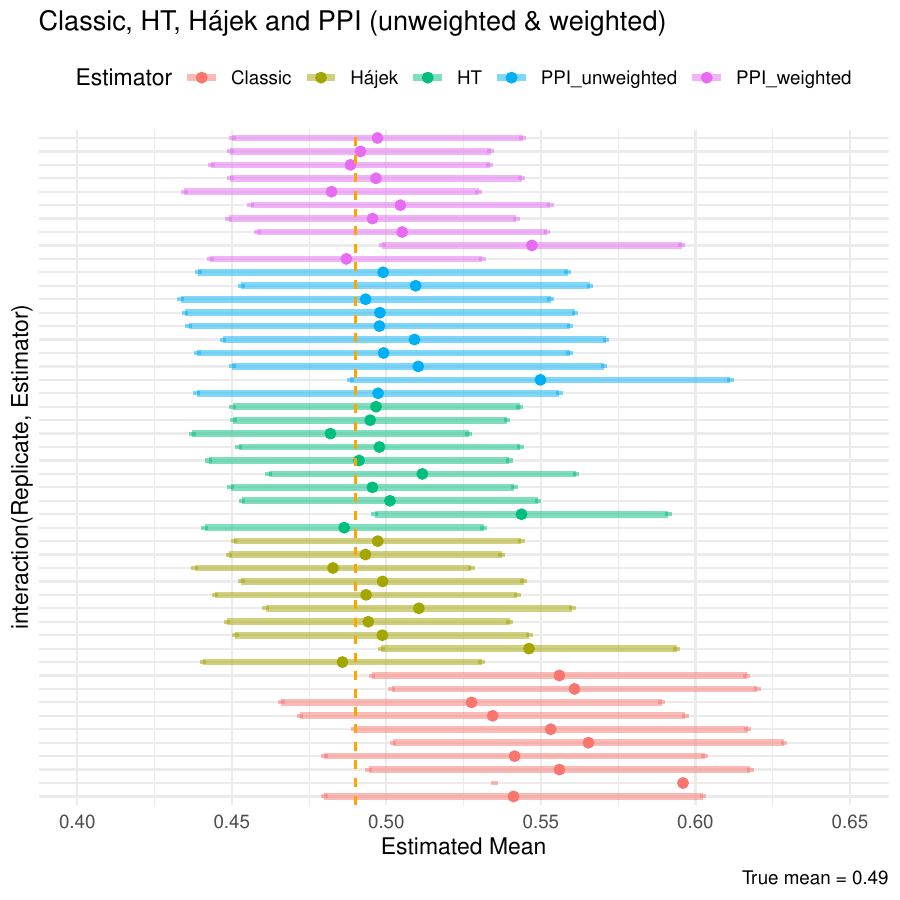}
    \caption{Horizontal 95\% confidence intervals for the first 10 replicates in the simulation with informative labeling with the dashed line indicating the true population mean.}
    \label{fig:placeholder}
\end{figure}

\begin{table}[ht]
\centering
\footnotesize{
\caption{Simulation results for mean estimation under informative labeling with estimated inclusion probabilities. Reported are the average point estimate, empirical bias, mean 95\% confidence interval width, and empirical coverage over 200 replicates. The true value of the mean parameter is $0.49$.}
\label{tab:sim-results-estimated}
\begin{tabular}{lrrrr}
\hline
Estimator & Estimate & Bias & Width & Coverage \\
\hline
Classic          & 0.543 & 0.0529  & 0.123 & 0.660 \\
Horvitz--Thompson & 0.491 & 0.00120 & 0.092 & 0.965 \\
H\'ajek          & 0.491 & 0.00118 & 0.092 & 0.965 \\
PPI (unweighted) & 0.498 & 0.00769 & 0.120 & 1.000 \\
PPI (weighted)   & 0.491 & 0.00119 & 0.092 & 0.975 \\
\hline
\end{tabular}}
\end{table}

Table~\ref{tab:sim-results-estimated} and Figure~\ref{fig:placeholder} summarize the performance of the candidate estimators under informative labeling with estimated propensities. The classical labeled-only estimator shows substantial bias (0.053) and severe undercoverage (66\%), confirming the failure of naive inference under informative labeling.

All IPW-corrected methods: Horvitz--Thompson, H\'ajek, and both PPI variants exhibit negligible bias and achieve near-nominal or conservative coverage. The HT and H\'ajek estimators achieve the narrowest confidence intervals (width 0.092) with excellent coverage (96.5\%), demonstrating the efficiency of ratio estimation when inclusion probabilities vary substantially. The weighted PPI approach matches this performance exactly in terms of both bias and interval width, achieving coverage of 97.5\%. The unweighted PPI estimator, while still substantially better than the classical approach, shows slightly higher bias (0.0077) and wider intervals (0.120) compared to the IPW-corrected methods. This difference highlights the value of weighting by inverse propensities when labeling is highly informative. Overall, the H\'ajek estimator and weighted PPI perform nearly identically in this setting, both leveraging ratio estimation to achieve efficient, unbiased inference.
\subsection{Additional Simulation Study: effect of $p_{\text{lab}}$}
\begin{figure*}[ht!]
    \centering
    \includegraphics[width=\linewidth]{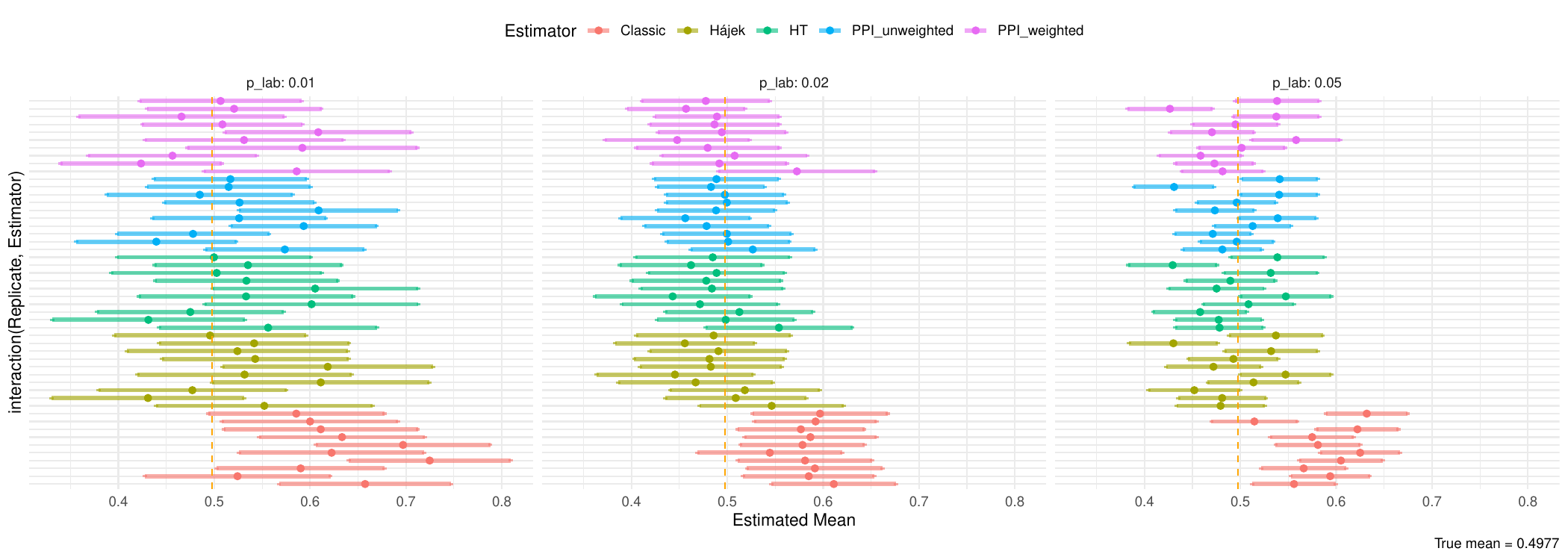}
    \caption{Horizontal 95\% confidence intervals for the first 10 replicates in the simulation with informative labeling, shown separately for labeled proportions $p_{\text{lab}}=0.01$, $0.02$, and $0.05$. The dashed line indicates the true population mean.}
    \label{fig:facet_plot}
\end{figure*}

Here we present additional simulation results under an informative labeling mechanism, for different labeled proportions $p_{\text{lab}} \in \{0.01, 0.02, 0.05\}$. The finite population size is $N = 10{,}000$, substantially larger than the labeled sample, creating a realistic semi-supervised setting. In each case, the finite population mean is the target estimand and the labeling probability depends on covariates, inducing bias in unweighted estimators. Table~\ref{tab:facet} reports the average estimate, bias, mean 95\% confidence interval (CI) width, empirical coverage, and the average number of labeled units across 200 replicates. Figure~\ref{fig:facet_plot} displays horizontal 95\% CIs for the first 10 replicates in each $p_{\text{lab}}$ setting.

\begin{table*}[ht!]
\centering
\caption{Simulation results for different labeled proportions $p_{\text{lab}}$ with $N=10{,}000$. 
Mean\_Estimate, Bias, Mean\_Width, Coverage, and average number of labeled units (Avg\_n\_lab) are computed over 200 replicates. The true population mean is $0.4977$.}
\label{tab:facet}
\footnotesize{
\setlength{\tabcolsep}{6pt}
\begin{tabular}{c l c c c c c}
\toprule
$p_{\text{lab}}$ & Estimator & Mean\_Estimate & Bias & Mean\_Width & Coverage & Avg\_n\_lab \\
\hline
0.01 & Classic         & 0.602 &  0.104 & 0.192 & 0.410 &  100 \\
0.01 & HT              & 0.500 &  0.003 & 0.219 & 0.975 &  100 \\
0.01 & H\'ajek         & 0.502 &  0.004 & 0.219 & 0.955 &  100 \\
0.01 & PPI\_unweighted & 0.505 &  0.008 & 0.178 & 0.965 &  100 \\
0.01 & PPI\_weighted   & 0.501 &  0.003 & 0.201 & 0.930 &  100 \\
\hline 
0.02 & Classic         & 0.597 &  0.099 & 0.136 & 0.190 &  200 \\
0.02 & HT              & 0.498 &  0.001 & 0.155 & 0.955 &  200 \\
0.02 & H\'ajek         & 0.499 &  0.001 & 0.155 & 0.960 &  200 \\
0.02 & PPI\_unweighted & 0.502 &  0.004 & 0.127 & 0.965 &  200 \\
0.02 & PPI\_weighted   & 0.498 &  0.000 & 0.142 & 0.955 &  200 \\
\hline
0.05 & Classic         & 0.595 &  0.097 & 0.086 & 0.005 &  500 \\
0.05 & HT              & 0.499 &  0.001 & 0.096 & 0.950 &  500 \\
0.05 & H\'ajek         & 0.499 &  0.002 & 0.096 & 0.960 &  500 \\
0.05 & PPI\_unweighted & 0.502 &  0.005 & 0.081 & 0.945 &  500 \\
0.05 & PPI\_weighted   & 0.498 &  0.001 & 0.089 & 0.950 &  500 \\
\hline
\end{tabular}}
\end{table*}

As expected, the classic estimator exhibits severe bias when labeling is informative, achieving poor coverage across all settings (41\%, 19\%, and 0.5\% for $p_{\text{lab}} \in \{0.01, 0.02, 0.05\}$, respectively). In contrast, all IPW-corrected methods achieve near-nominal coverage (93--98\%) across all labeled fractions.

Table~\ref{tab:facet} demonstrates that the weighted PPI estimator consistently achieves the lowest bias among all methods. At $p_{\text{lab}} = 0.01$, weighted PPI has bias of only 0.003 compared to 0.008 for unweighted PPI. At $p_{\text{lab}} = 0.02$, weighted PPI shows negligible bias while unweighted PPI exhibits bias of 0.004. This superior bias performance continues at $p_{\text{lab}} = 0.05$, where weighted PPI maintains bias of 0.001 compared to 0.005 for unweighted PPI. The classical ratio estimators HT and H\'ajek show intermediate bias performance, generally between the two PPI variants.

All IPW-corrected methods achieve similar coverage rates (93--98\%) and comparable interval widths, with no method uniformly dominating in terms of width. At small labeled fractions ($p_{\text{lab}} = 0.01$), unweighted PPI achieves narrower intervals (0.178) than weighted PPI (0.201), but this advantage comes at the cost of higher bias. The classical H\'ajek and HT estimators produce identical interval widths at each $p_{\text{lab}}$ level. As the labeled fraction increases, all methods show substantial reductions in interval width as expected, with widths decreasing from around 0.18--0.22 at $p_{\text{lab}} = 0.01$ to 0.08--0.10 at $p_{\text{lab}} = 0.05$.

Overall, the weighted PPI estimator offers the best bias-variance tradeoff, achieving consistently lower bias than competing methods while maintaining competitive interval width and near-nominal coverage across all labeling scenarios.

\section{Discussion}\label{sec:diss}

We show that inverse probability weighting can be used within the prediction-powered inference framework to handle informative labeling, with the Horvitz--Thompson and H\'ajek forms providing natural bias-correcting rectifiers. This connection brings together design-based survey sampling ideas and modern prediction-assisted inference, yielding estimators that remain valid when labeling probabilities vary across units. 

An important direction for future work is the development of rigorous variance expressions and confidence interval procedures for cross-PPI estimators under covariate shift and for binning-smoothing variants of the Horvitz--Thompson estimator. While our empirical results demonstrate good finite-sample performance, formal theoretical guarantees would strengthen the foundations of these methods.

Our simulations also provide a robustness check: when the inclusion probabilities are not known but are estimated from a correctly specified model, performance is essentially unchanged from the known-\(\xi_i\) case. This suggests that the practical applicability of IPW-adjusted PPI could be broader than the idealized setting hitherto considered.

\bibliographystyle{imsart-nameyear} % Style BST file (imsart-number.bst or imsart-nameyear.bst)
%\bibliography{bibliography}       % Bibliography file (usually '*.bib')

% \bibliographystyle{apalike} 
\bibliography{references,ppi_refs}
%% or include bibliography directly:
% \begin{thebibliography}{}
% \bibitem{b1}
% \end{thebibliography}

\end{document}